\documentclass{egpubl}
\usepackage{pg2026s}

\pdfpagewidth=\paperwidth
\pdfpageheight=\paperheight

\WsConferencePaper

\usepackage[T1]{fontenc}
\usepackage{dfadobe}

\usepackage{cite}  
\BibtexOrBiblatex
\electronicVersion
\PrintedOrElectronic
\hypersetup{
  hypertexnames=false,
  colorlinks=true,
  linkcolor=black,
  citecolor=black,
  urlcolor=black,
  pdfsubject={Pacific Graphics 2026, Conference Paper},
  pdfkeywords={Motion capture, 3D human reconstruction, character animation, monocular video, camera-motion disentanglement}
}
\ifpdf \usepackage[pdftex]{graphicx} \pdfcompresslevel=9
\else \usepackage[dvips]{graphicx} \fi

\usepackage{egweblnk}

\usepackage{multirow}
\usepackage{multicol}
\usepackage{makecell}
\usepackage{booktabs}
\usepackage{amsmath}
\usepackage{amssymb}
\usepackage{bm}
\usepackage{xcolor}
\usepackage{float}
\newcommand{\rev}[1]{#1}
\title[WATCH: Allied Trajectory and Pose for Camera and Human]%
      {WATCH: World-aware Allied Trajectory and pose reConstruction for Camera and Human}

\author[Z. Hu et al.]%
{Zhongyuan Hu$^{1,2,*}$,
 Qijun Ying$^{2,*}$,
 Jiazhi Shu$^{3}$,
 Ronghui Li$^{1}$,
 Yu Lu$^{2}$,
 Zijiao Zeng$^{2,\dagger}$, and
 Xiu Li$^{4,\dagger}$\\
$^{1}$Tsinghua University; $^{2}$Tencent\\
$^{3}$The Hong Kong University of Science and Technology; $^{4}$TSIG, Tsinghua University\\
$^{*}$Equal contribution. $^{\dagger}$Corresponding authors:
\texttt{zijiaozeng@tencent.com}; \texttt{li.xiu@sz.tsinghua.edu.cn}}

\begin{document}

\teaser{
  \includegraphics[width=\linewidth]{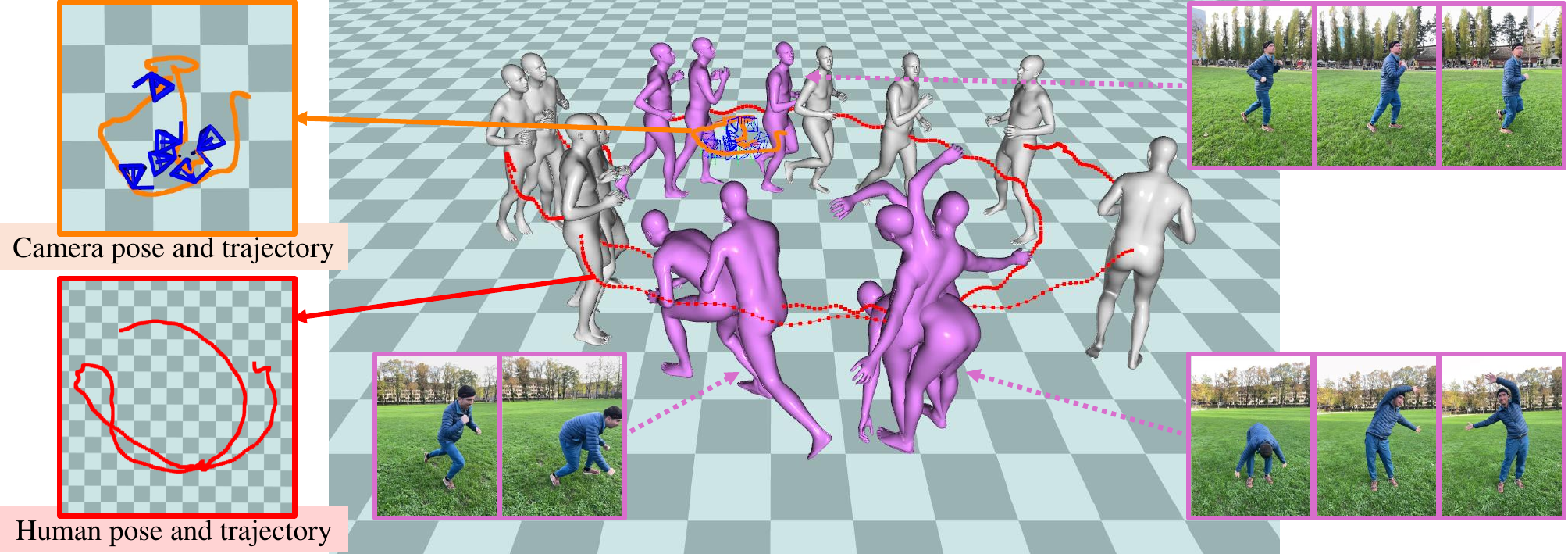}
  \centering
  \caption{Left: Model-predicted top-down view of camera poses and trajectories~(top) alongside human poses and trajectories~(bottom).
           Right: Global trajectory reconstruction with corresponding input frames.
           Unlike previous methods that focus solely on human motion, WATCH simultaneously reconstructs camera and human motions,
           enabling better bridging between camera-space and world-space cues for improved trajectory accuracy.}
  \label{fig:teaser}
}

\maketitle
\hypersetup{
  pdftitle={WATCH: World-aware Allied Trajectory and pose reConstruction for Camera and Human},
  pdfauthor={Zhongyuan Hu, Qijun Ying, Jiazhi Shu, Ronghui Li, Yu Lu, Zijiao Zeng, and Xiu Li}
}
\begin{abstract}
Reconstructing global human motion from monocular video is fundamental to VR, graphics, and robotics, yet remains ill-posed due to depth ambiguity, motion ambiguity, and the entanglement of camera and human movements. Human-motion-centric methods achieve strong physical plausibility but leave two signals unused: camera orientation is processed through a fixed coordinate transformation with no independent supervision of its components, and camera velocity is discarded entirely despite being directly observable from SLAM. Camera-trajectory-centric methods use camera translation directly, but hard-decoding SLAM trajectories into human positions propagates depth errors and fails entirely under static cameras. We present \textbf{WATCH} (World-aware Allied Trajectory and pose reConstruction for Camera and Human). The key observation is that once camera orientation is made explicit, camera velocity becomes a natural additional input rather than an ambiguous one. We therefore decompose camera rotation into a network-estimated roll-pitch component and an analytically recoverable yaw, supervising each independently. This decomposition exposes a clean geometric interface through which camera velocity is incorporated as a learned spatial prior in the backbone, without the physically implausible artifacts that arise from hard-decoding. WATCH outperforms prior human-motion-centric methods on both static-camera (RICH) and dynamic-camera (EMDB) benchmarks in global trajectory accuracy, temporal smoothness, and physical plausibility, and remains robust when ground-truth camera is replaced with DPVO estimates.


\vspace{-4pt}

\end{abstract}
\section{Introduction}
\label{sec:intro}

Global human motion reconstruction from monocular video is fundamental to VR~\cite{shrestha2024generating}, graphics~\cite{li2025genmo,han2024atomaligningtexttomotionmodel,li2026infinitedancescalable3ddance,ma2026controllablevideogenerationsurvey}, and robotics~\cite{fu2024humanplus}. Unlike standard human pose estimation, this task requires estimating motion in world coordinates, which demands accurate decoupling of camera and human motion under depth ambiguity and scale uncertainty.

With the advancement of SLAM~\cite{droid, dpvo, orb2, wang2024dust3rgeometric3dvision}, two paradigms have emerged for world-coordinate reconstruction. \textbf{Camera-trajectory-centric methods}~\cite{tram,whac,human3r} map SLAM trajectories directly to human positions. This hard-decoding achieves strong performance on dynamic-camera sequences but collapses under static cameras where SLAM provides no translational signal, and propagates depth errors directly into pose estimates. \textbf{Human-motion-centric methods}~\cite{wham,gvhmr} instead estimate human velocity from visual cues, achieving better physical plausibility and generalization. However, they leave two informative signals unused. First, camera orientation is processed through a fixed coordinate transformation with no independent supervision of roll, pitch, or yaw---so orientation errors cannot be explicitly corrected. Second, camera translation velocity, which is directly observable from SLAM, is discarded entirely.

The two limitations are linked. Without an explicit orientation decomposition, injecting raw camera translation into the network creates ambiguous feature interactions: the model cannot distinguish between translation-induced and motion-induced position changes. Solving orientation first therefore unlocks translation.

We present \textbf{WATCH} (\textbf{W}orld-aware \textbf{A}llied \textbf{T}rajectory and pose re\textbf{C}onstruction for Camera and \textbf{H}uman), which addresses both issues through two complementary designs.

\textbf{Analytical heading decomposition.} We decompose camera rotation into a network-estimated roll-pitch component and an analytically recoverable yaw, supervising each independently. Unlike GVHMR's view-gravity operator---a fixed transform applied at output with no per-component gradient---our decomposition exposes roll-pitch as a direct regression target and recovers yaw analytically from the estimated roll-pitch and supplied relative camera rotation. This creates a clean geometric interface through which camera translation can enter the network without orientation ambiguity.

\textbf{Soft camera trajectory conditioning.} Rather than hard-decoding SLAM output into human positions as in TRAM~\cite{tram} and WHAC~\cite{whac}, we encode camera velocity as a learned spatial prior in the backbone. The network learns the statistical relationship between camera and human motion from training data, preserving physical plausibility while exploiting translation cues that prior human-motion-centric methods ignore. The main contributions of this work are:

\begin{enumerate}
    \item We identify that implicit orientation handling in human-motion-centric methods both limits orientation accuracy and prevents camera translation from being used. Our analytical heading decomposition supervises roll-pitch and yaw independently, resolving both issues with a single design.
    \item We introduce soft camera trajectory conditioning, the first mechanism to incorporate camera translation velocity as a learned prior in a human-motion-centric framework, without hard-decoding SLAM output.
    \item WATCH achieves state-of-the-art performance among human-motion-centric methods on both static-camera (RICH) and dynamic-camera (EMDB) benchmarks, demonstrating generalization that camera-trajectory-centric methods cannot achieve when camera motion is absent.
\end{enumerate}

\section{Related Work}

\noindent\textbf{Human Pose and Shape Estimation~(HPS).} Human motion recovery from monocular video aims to estimate 3D pose and shape in camera coordinates using parametric models such as SMPL~\cite{smpl} and SMPL-X~\cite{smplx}. Early optimization-based methods~\cite{opt_smplify} minimized reprojection errors for pose fitting. Regression-based methods~\cite{hmr, spin, hybrik, pymaf} established deep learning as the dominant paradigm. CLIFF~\cite{cliff} showed that modeling perspective projection is essential for spatial plausibility, motivating subsequent work~\cite{refit, camerahmr}. HMR2.0~\cite{hmr2} applied pretrained ViTs to achieve strong single-frame performance. Temporal models~\cite{vibe, meva, tcmr, glot, maed} extended these baselines with RNNs and Transformers to reduce jitter and handle occlusion, but most approaches operate in camera coordinates and neglect world-space motion entirely.

\begin{figure*}[!htbp]
  \centering
  \includegraphics[width=1\textwidth]{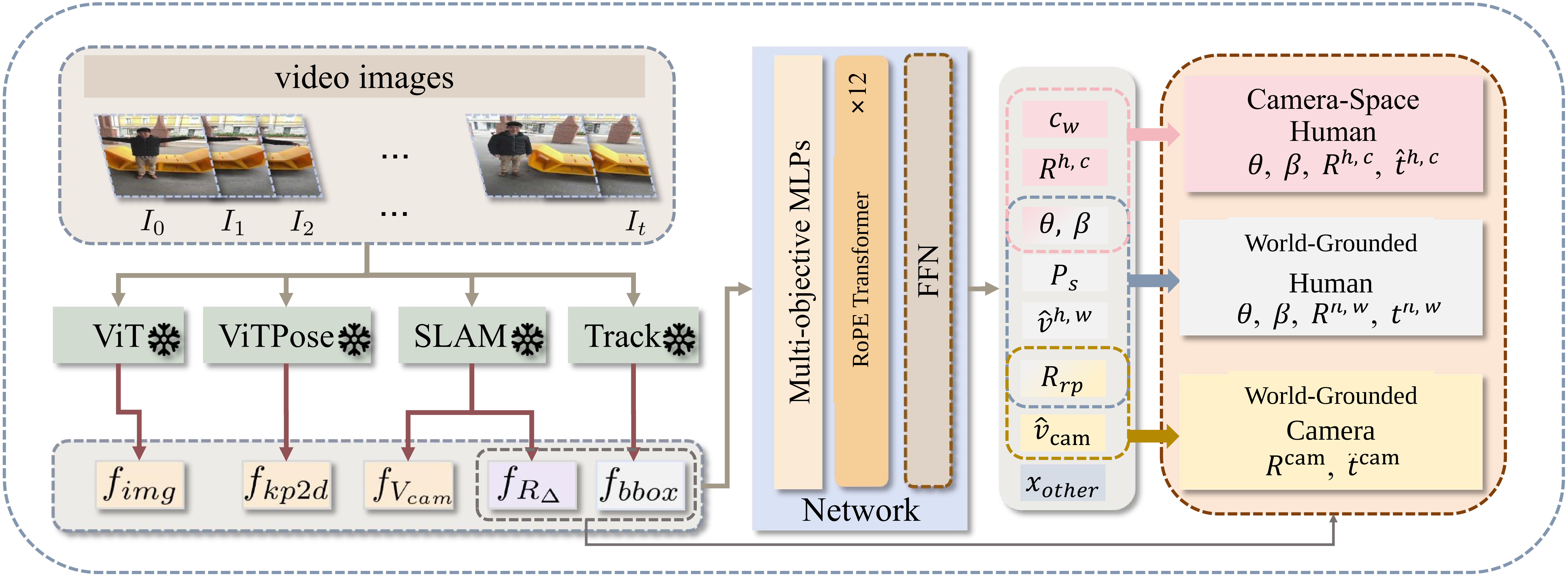}
  \caption{ \textbf{Overview of the WATCH framework.}
    Given a monocular video sequence, WATCH extracts multi-modal inputs: ViT-based image features $f_{img}$, human bounding boxes $f_{bbox}$, 2D keypoints $f_{kp2d}$ from ViTPose, camera rotation angles $f_{R_\Delta}$, and camera local velocities $f_{V_{cam}}$. These are fused and processed by a RoPE Transformer Encoder that jointly predicts camera-space human poses, camera roll-pitch angles, human local velocity, and camera local velocity. The predicted roll-pitch component and camera angular velocity feed into our analytical heading decomposition~(Sec.~\ref{sec:heading_decomposition}) to recover world-space human and camera orientations, while the velocity predictions are integrated into full trajectories~(Sec.~\ref{sec:trajectory_integration}). This unified design enables end-to-end optimization of both camera and human motion in a single forward pass.
    }
  \label{fig:architecture}
\end{figure*}

\noindent\textbf{From Camera Space to World Space.} World-coordinate human motion reconstruction requires estimating human movement relative to the physical world rather than camera coordinates, involving complex decoupling of camera and human motions. Early optimization-based methods~\cite{physcap, rfc} attempted to reconstruct global motion from monocular videos through physical constraints, but suffered from high computational complexity and sensitivity to initialization. Subsequent hybrid approaches~\cite{d&d, physpt} fed HPS predictions into physics simulators for optimization, improving physical plausibility but heavily depending on HPS prediction quality and struggling with complex outdoor scenarios.

With the maturation of SLAM technology, environment reconstruction methods~\cite{slahmr, synchmr, tram} and human-scene constraint approaches~\cite{josh} emerged, providing important contextual information but heavily relying on SLAM reconstruction quality and camera-space prediction stability, leading to practical application challenges.

Recent advances have been led by WHAM~\cite{wham}, which demonstrates remarkable plausibility and reliability in human motion reconstruction, both in camera space and world space. WHAM integrates ViT image features and camera angular velocity through a simple RNN architecture, gaining widespread community adoption. Building upon this foundation, GVHMR~\cite{gvhmr} introduced an elegant mechanism reminiscent of CLIFF's approach: the view-gravity operator implements heading angle decomposition to successfully mitigate error accumulation in human orientation estimation. HumanMM~\cite{zhang2025humanmmglobalhumanmotion} further expanded the capability boundaries of in-the-wild motion capture models, ensuring motion consistency across multi-camera videos from a human motion perspective.

These observations raise a fundamental question: rather than mapping camera trajectories directly to human positions, can camera translation serve as a \emph{learned spatial prior} within the reconstruction network itself? This reframing motivates WATCH, which treats camera translation velocity not as a hard geometric constraint but as contextual input that guides the model's internal spatial reasoning.

\noindent\textbf{Joint Human-Scene Reconstruction.} Recent work~\cite{josh, human3r, wang2025vggtvisualgeometrygrounded} jointly reasons about cameras, humans, and scene geometry through optimization or feed-forward reconstruction. This richer setting is complementary to WATCH's operating point: WATCH avoids dense scene reconstruction and uses camera translation only as a lightweight learned condition. Its estimates could initialize joint optimization, while the supervised orientation interface could also be used as an objective in scene-aware systems.

\section{Methodology}
Figure~\ref{fig:architecture} shows the WATCH framework. It has two core components: an analytical heading decomposition with supervised roll-pitch and analytically recovered yaw, and a camera trajectory integration that encodes camera velocity as a learned spatial prior in the backbone.

\subsection{Problem Formulation}
\label{sec:formulation}

We represent 3D human motion using SMPL-X~\cite{smplx} as $\mathcal{M}(\theta, \beta, \mathbf{R}^h, \mathbf{t}^h)$, where $\theta \in \mathbb{R}^{21 \times 3}$ encodes body joint rotations, $\beta \in \mathbb{R}^{10}$ describes body shape, and $\mathbf{R}^h \in SO(3)$, $\mathbf{t}^h \in \mathbb{R}^3$ are the root rotation and translation. Given a monocular video $\{I_t\}_{t=1}^T$ and camera extrinsics $\{\mathbf{R}_t^{cam}, \mathbf{t}_t^{cam}\}_{t=1}^T$, we estimate pose parameters $\{\theta_t, \beta_t\}$, camera-space human motion $\{\mathbf{R}_t^{h,c}, \mathbf{t}_t^{h,c}\}$, and world-space human trajectory $\{\mathbf{R}_t^{h,w}, \mathbf{t}_t^{h,w}\}$.

\begin{figure}[!ht]
  \centering
  \includegraphics[width=\linewidth]{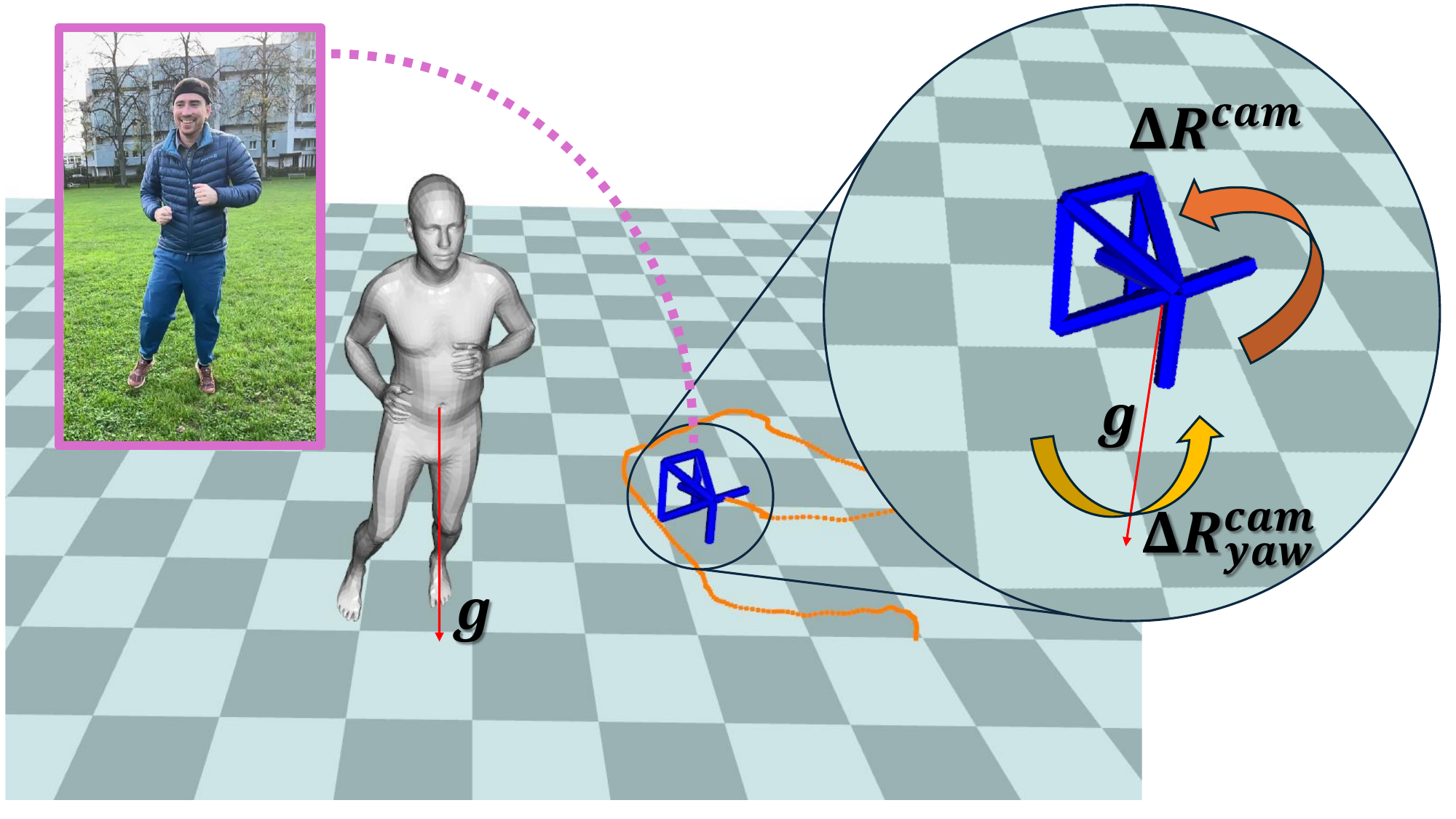}
  \caption{\textbf{Illustration of analytical heading decomposition.} The full camera rotation $\mathbf{R}^{cam}$ is decomposed into a roll-pitch component $\mathbf{R}_{rp}^{cam}$, which aligns the camera with the gravity axis $g$, and a heading (yaw) component $\mathbf{R}_{yaw}^{cam}$, which captures rotation around $g$. The camera angular velocity $\Delta\mathbf{R}^{cam}$ is expressed in the body frame, allowing the heading angular velocity $\Delta\mathbf{R}_{yaw}^{cam}$ to be recovered analytically from the network-estimated roll-pitch angles and supplied relative camera rotation~(Eq.~\ref{eq:yaw_velocity}). This decomposition eliminates the need to directly regress the full 3D orientation, reducing the estimation problem to the more tractable roll-pitch subspace.}
  \label{fig:angle}
\end{figure}

\subsection{Analytical Heading Decomposition}
\label{sec:heading_decomposition}

GVHMR~\cite{gvhmr} applies a view-gravity coordinate transform only at its output, whereas WATCH exposes roll-pitch as an explicit regression target and analytically recovers relative heading. This provides a well-defined orientation state in which local camera translation can be injected without coupling it to an implicitly estimated yaw. WATCH also predicts camera motion as an auxiliary task.

To isolate the heading component of camera rotation, we employ rotation decomposition in the world reference frame. Using fixed-axis Euler decomposition, we decompose the camera rotation matrix as:
\begin{equation}
\mathbf{R}_t^{cam} = \mathbf{R}_{yaw,t}^{cam} \mathbf{R}_{rp,t}^{cam}
\label{eq:cam_decomp}
\end{equation}
where $\mathbf{R}_{rp,t}^{cam}$ represents the roll-pitch component and $\mathbf{R}_{yaw,t}^{cam}$ denotes the heading component.

The heading angular velocity $\Delta\mathbf{R}_{yaw,t}^{cam} = (\mathbf{R}_{yaw,t}^{cam})^T \mathbf{R}_{yaw,t+1}^{cam}$ can be obtained through:
\begin{equation}
\Delta\mathbf{R}_{yaw,t}^{cam} = \mathbf{R}_{rp,t}^{cam} \Delta\mathbf{R}_t^{cam} (\mathbf{R}_{rp,t+1}^{cam})^T
\label{eq:yaw_velocity}
\end{equation}
where $\Delta\mathbf{R}_t^{cam} = (\mathbf{R}_t^{cam})^T \mathbf{R}_{t+1}^{cam}$ is body-frame camera angular velocity and $\mathbf{R}^{cam}$ denotes camera-to-world rotation. When a camera front end supplies the equivalent world-to-camera increment, we transpose it before applying Eq.~\ref{eq:yaw_velocity}. In the body frame, the roll-pitch component acts as a basis change that aligns angular velocity with gravity; heading then becomes a pure rotation about gravity. World-frame angular velocity instead couples heading and tilt non-linearly.

Given this decomposition, the network only needs to estimate the camera's roll-pitch component $\mathbf{R}_{rp,t}^{cam}$. The heading orientation is obtained through recursive integration:
\begin{equation}
\mathbf{R}_{yaw,t}^{cam} = \mathbf{R}_{yaw,0}^{cam} \prod_{i=0}^{t-1} \Delta\mathbf{R}_{yaw,i}^{cam}
\label{eq:yaw_integration}
\end{equation}

This formulation is invariant to the initial heading orientation: world-space poses depend only on relative changes $\Delta\mathbf{R}_{yaw}^{cam}$, so the network focuses solely on the roll-pitch component. During training, $\mathbf{R}_{yaw,0}^{cam}$ is set to a canonical reference; during inference it is simply the identity matrix. Combined with the network-estimated camera-space human orientation $\mathbf{R}_t^{h,c}$, we obtain the human orientation in world space:
\begin{equation}
\mathbf{R}_t^{h,w} = \mathbf{R}_{yaw,t}^{cam} \mathbf{R}_{rp,t}^{cam} \mathbf{R}_t^{h,c}
\label{eq:human_world_orient}
\end{equation}

Through this analytical approach, estimating only the camera's roll-pitch component and velocity suffices to determine both camera and human orientations in world space. This significantly simplifies the estimation task compared to existing implicit methods while providing a foundation for algorithmic extensions that integrate camera motion information into the learning framework.

\newcolumntype{?}{!{\vrule width 0.75pt}}

\begin{table*}[tbh]
    \centering
    \vspace{-4mm}
    \setlength{\tabcolsep}{2pt}
    \renewcommand{\arraystretch}{1.05}
    \resizebox{\textwidth}{!}
    {\small{
        \begin{tabular}{l l?ccccc?ccccc}
            \cmidrule[0.75pt]{1-12}
            & & \multicolumn{5}{c}{RICH (24)} & \multicolumn{5}{c}{EMDB-2 (24)} \\
            \cmidrule(lr){3-7} \cmidrule(lr){8-12}
            & Models & \scriptsize{WA-MPJPE$_{100}$}$\downarrow$ & \scriptsize{W-MPJPE$_{100}$}$\downarrow$ & \scriptsize{RTE}$\downarrow$ & \scriptsize{Jitter}$\downarrow$ & \scriptsize{Foot-Sliding}$\downarrow$ & \scriptsize{WA-MPJPE$_{100}$}$\downarrow$ & \scriptsize{W-MPJPE$_{100}$}$\downarrow$ & \scriptsize{RTE}$\downarrow$ & \scriptsize{Jitter}$\downarrow$ & \scriptsize{Foot-Sliding}$\downarrow$ \\
            \cmidrule{1-12}


            \multicolumn{2}{l}{\hspace{-2pt}\textbf{Camera-trajectory-centric}} \\

            & SLAHMR~\cite{slahmr} & \textbf{98.1} & \underline{186.4} & 28.9 & -- & -- & 326.9 & 776.1 & 10.2 & 31.3 & \underline{14.5} \\
            & WHAC~\cite{whac} & -- & -- & -- & -- & -- & 142.2 & 343.3 & -- & -- & -- \\

            & TRAM~\cite{tram} & 127.8 & 238.0 & \underline{6.0} & -- & -- & \underline{76.4} & \underline{222.4} & \underline{1.4} & \underline{18.5} & 23.4 \\
            & Human3R~\cite{human3r} & \underline{110.0} & \textbf{184.9} & \textbf{3.3} & -- & -- & 112.2 & 267.9 & 2.2 & -- & -- \\
            & PromptHMR-vid~\cite{wang2025prompthmrpromptablehumanmesh} & -- & -- & -- & -- & -- & \textbf{71.0} & \textbf{216.5} & \textbf{1.3} & \textbf{16.3} & \textbf{3.5} \\
            \cmidrule[0.75pt]{1-12}
            \multicolumn{2}{l}{\hspace{-2pt}\textbf{Human-motion-centric}} \\

            & GLAMR~\cite{glamr} & 129.4 & 236.2 & 3.8 & 49.7 & 18.1 & 280.8 & 726.6 & 11.4 & 46.3 & 20.7 \\
            & TRACE~\cite{trace} & 238.1 & 925.4 & 610.4 & 1578.6 & 230.7 & 529.0 & 1702.3 & 17.7 & 2987.6 & 370.7 \\
            & WHAM~\cite{wham} & 109.9 & 184.6 & 4.1 & 19.7 & 3.3 & 135.6 & 354.8 & 6.0 & 22.5 & 4.4 \\
            & {GVHMR}~\cite{gvhmr} & \underline{78.8} & \underline{126.3} & \underline{2.4} & \underline{12.8} & \underline{3.0} & 111.0 & 276.5 & 2.0 & 16.7 & \underline{3.5} \\
            & WATCH~(EMDB: DPVO, Ours) & \textbf{66.8} & \textbf{107.9} & \textbf{2.1} & \textbf{5.3} & \textbf{2.2} & \underline{105.2} & \underline{256.9} & \underline{1.9} & \underline{10.4} & \textbf{3.1} \\
            & WATCH~(EMDB: GT cam, Ours) & \textbf{66.8} & \textbf{107.9} & \textbf{2.1} & \textbf{5.3} & \textbf{2.2} & \textbf{103.3} & \textbf{251.6} & \textbf{1.7} & \textbf{9.8} & \textbf{3.1} \\
            \cmidrule[0.75pt]{1-12}
        \end{tabular}
    }}
    \caption{
    \textbf{World-grounded metrics.} Parentheses denote the evaluated joint count. RICH uses its provided static camera; its WATCH values are repeated only for side-by-side comparison of the two EMDB-2 camera sources. Bold and underline denote numerically best and second-best values within each method block under each method's published protocol, \rev{not a common-protocol ranking}. PromptHMR-vid combines its predictor with TRAM metric SLAM.
    }
    \label{tab:quant_global}
\end{table*}

\subsection{Camera Trajectory Integration}
\label{sec:trajectory_integration}

Hard-decoding camera positions into human poses fails because SLAM drift and depth ambiguity propagate directly into pose estimates. We instead encode camera velocity as spatial contextual input in the embedding space, supervised through an auxiliary trajectory prediction task.

Each frame token is initialized from 17 confidence-weighted 2D joint observations. Separate modality projectors map the HMR2.0 image feature, CLIFF bounding-box/intrinsics feature, normalized body-frame camera-angular-velocity feature, and local camera-translation velocity to the same latent width. \rev{Specifically, $\mathbf{f}_{2D,t}\equiv\mathbf{f}_{kp2d,t}$ is the projected 2D-keypoint feature shown in Fig.~\ref{fig:architecture}, $\mathbf{f}_{V_{cam},t}=P_V(\mathbf{v}_t^{cam})$ is the projected local camera-translation velocity, and $\mathcal{F}_{others,t}=\{\mathbf{f}_{img,t},\mathbf{f}_{bbox,t},\mathbf{f}_{\Delta R,t}\}$ contains the remaining projected conditions.} Here $\mathbf{f}_{\Delta R,t}$ is the projected normalized body-frame camera-angular-velocity feature. We fuse them by addition:
\begin{equation}
\mathbf{f}_t = \mathbf{f}_{2D,t} + \mathbf{f}_{V_{cam},t}
                 + \sum_{\mathbf{f}\in\mathcal{F}_{others,t}}\mathbf{f}
\label{eq:feature_fusion}
\end{equation}
After the backbone Transformer, separate heads predict human local velocity $\hat{\mathbf{v}}_t^h$ and camera local velocity $\hat{\mathbf{v}}_t^{cam}$. \rev{Their trajectories are integrated without an ambiguous shared velocity symbol:}
\begin{equation}
\hat{\mathbf{t}}_t^q = \mathbf{t}_0^q
  + \sum_{i=1}^t \mathbf{R}_i^q\hat{\mathbf{v}}_i^q,
\qquad q\in\{h,cam\}.
\label{eq:trajectory_integration}
\end{equation}

\subsection{Model Outputs and Predictions}
\label{sec:model_outputs}

WATCH predicts three groups of outputs: camera-space human poses, world-space human motion, and camera poses with trajectories.

\textbf{Camera-space Human Reconstruction}: The model predicts standard SMPL parameters including body pose $\theta$ and shape $\beta$, along with camera-space human orientation $\mathbf{R}^{h,c}$ and CLIFF~\cite{cliff} weak-perspective camera parameters $c_w$. These outputs ensure compatibility with existing HPS pipelines while providing the foundation for world-space transformation.

\textbf{World-space Human Motion}: For world-space human motion, we estimate several key components: camera roll-pitch angles $\mathbf{R}_{rp}^{cam}$, contact estimates for hands and feet to maintain physical plausibility~$P_s$, and human local velocity $\hat{\mathbf{v}}^{h,w}$ in the body coordinate system. The human orientation in world space $\mathbf{R}^{h,w}$ is derived analytically through our heading decomposition formulation (Equation~\ref{eq:human_world_orient}), eliminating the need for direct regression of this complex quantity.

\textbf{Camera Motion Estimation}: Camera trajectories are reconstructed using the predicted roll-pitch angles and estimated camera velocity $\hat{\mathbf{v}}^{cam}$. This auxiliary task enforces spatial consistency between human and camera motion during training. The model additionally predicts heading angular velocities for both human and camera as auxiliary supervision signals.

\subsection{Loss Function Design}
\label{sec:loss_function}

The submitted Full WATCH checkpoint and the corresponding Full WATCH results use the following implementation-authoritative objective:
\begin{equation}
\mathcal{L} = \mathcal{L}_{simple} + \mathcal{L}_{hmr}
 + \mathcal{L}_{traj}^h + \mathcal{L}_{traj}^{cam}
 + \mathcal{L}_{static} + \lambda_{accel}\mathcal{L}_{accel}.
\label{eq:total_loss}
\end{equation}
The acceleration-loss weight is \rev{$\lambda_{accel}=30$}.
The \rev{L2 trajectory notation and $\lambda_{cam}=0.25$} in the submitted main paper, and the \rev{omission of $\mathcal{L}_{static}$ and $\mathcal{L}_{accel}$} from the submitted supplement, were documentation errors rather than different Full-model settings.

\textbf{Direct Prediction Loss} $\mathcal{L}_{simple}$: \rev{MSE} directly supervises the encoded HMR, human-motion, and camera-motion outputs.

\textbf{Camera-space Reconstruction Loss} $\mathcal{L}_{hmr}$: Standard HMR losses supervise root-relative 3D joints and vertices \rev{(weight 500 each)}, 2D joints and vertices \rev{(weight 1000 each)}, and the weak-perspective camera parameter \rev{(weight 1)}.

\textbf{Human Trajectory Consistency Loss} $\mathcal{L}_{traj}^h$: We employ two teacher forcing strategies to improve trajectory supervision. Let $\hat{\mathbf{v}}^h$ and $\hat{\mathbf{R}}^h$ denote the predicted human velocity and orientation, and $\mathbf{v}^h_{gt}$, $\mathbf{R}^h_{gt}$ their ground-truth counterparts. Two trajectory rollouts are computed via Equation~\ref{eq:trajectory_integration}:
\begin{equation}
\hat{\mathbf{t}}^{(1)}_t = \mathbf{t}_0 + \textstyle\sum_{i=1}^t \mathbf{R}^h_{gt,i}\,\hat{\mathbf{v}}^h_i, \quad
\hat{\mathbf{t}}^{(2)}_t = \mathbf{t}_0 + \textstyle\sum_{i=1}^t \hat{\mathbf{R}}^h_i\,\mathbf{v}^h_{gt,i}
\label{eq:teacher_forcing}
\end{equation}
The loss is the sum of \rev{L1} distances of both rollouts to the ground-truth trajectory $\mathbf{t}_{gt}$:
\rev{$\mathcal{L}_{traj}^h = \|\hat{\mathbf{t}}^{(1)} - \mathbf{t}_{gt}\|_1 + \|\hat{\mathbf{t}}^{(2)} - \mathbf{t}_{gt}\|_1$.}
This decoupled supervision allows orientation and velocity errors to be corrected independently, mitigating error accumulation during integration.

\textbf{Camera Motion Constraint Loss} $\mathcal{L}_{traj}^{cam}$: Identical \rev{L1} teacher-forced rollouts supervise predicted camera velocity and orientation. \rev{Both trajectory losses have unit weight.}

\textbf{Static-contact and Acceleration Losses}: $\mathcal{L}_{static}$ is \rev{unit-weight BCE on six automatically labelled hand/foot contact logits}. To suppress pose jitter, $\mathcal{L}_{accel}$ applies \rev{componentwise smooth-L1 with $\beta=0.02$} to second differences of predicted camera-coordinate root-relative joints:
\begin{align}
\mathbf{a}_t &={}
 \hat{\mathbf{J}}^{c,rel}_{t-1}-2\hat{\mathbf{J}}^{c,rel}_{t}
 +\hat{\mathbf{J}}^{c,rel}_{t+1},\nonumber\\
\mathcal{L}_{accel} &={}
 \operatorname*{mean}_{t,j,k\,\mathrm{valid}}\rho_{\beta}(a_{t,j,k}).
\label{eq:accel_loss}
\end{align}
\begin{equation}
\rho_{\beta}(x)=
\begin{cases}
x^2/(2\beta), & |x|<\beta,\\
|x|-\beta/2, & \text{otherwise},
\end{cases}
\end{equation}
This is averaged over valid frame triplets, joints, and Cartesian components.

\section{Experiment}
\begin{table*}[!tbh]
    \centering
    \vspace{-2mm}
    \setlength{\tabcolsep}{2pt}
    \renewcommand{\arraystretch}{1.05}
    \resizebox{\textwidth}{!}
    {\small{
        \begin{tabular}{cl?cccc?cccc?cccc}
            \cmidrule[0.75pt]{1-14}
            && \multicolumn{4}{c}{3DPW (14)} & \multicolumn{4}{c}{EMDB (24)} & \multicolumn{4}{c}{RICH (24)} \\
            \cmidrule(lr){3-6} \cmidrule(lr){7-10} \cmidrule(lr){11-14}
            & Models & \scriptsize{PA-MPJPE}$\downarrow$ & \scriptsize{MPJPE}$\downarrow$ & \scriptsize{PVE}$\downarrow$ & \scriptsize{Accel}$\downarrow$
            & \scriptsize{PA-MPJPE}$\downarrow$ & \scriptsize{MPJPE}$\downarrow$ & \scriptsize{PVE}$\downarrow$ & \scriptsize{Accel}$\downarrow$
            & \scriptsize{PA-MPJPE}$\downarrow$ & \scriptsize{MPJPE}$\downarrow$ & \scriptsize{PVE}$\downarrow$ & \scriptsize{Accel}$\downarrow$ \\
            \cmidrule{1-14}

            \multirow{10}{1em}{\rotatebox[origin=c]{90}{\textbf{per-frame}}}

            & SPIN~\cite{spin} & 59.2 & 96.9 & 112.8 & 31.4 & 87.1 & 140.3 & 174.9 & 41.3 & 69.7 & 122.9 & 144.2 & 35.2 \\
            & PARE$^*$~\cite{pare} & 46.5 & 74.5 & 88.6 & -- & 72.2 & 113.9 & 133.2 & -- & 60.7 & 109.2 & 123.5 & -- \\
            & CLIFF$^*$~\cite{cliff} & 43.0 & 69.0 & 81.2 & 22.5 & 68.1 & 103.3 & 128.0 & 24.5 & 56.6 & 102.6 & 115.0 & 22.4 \\
            & HybrIK$^*$~\cite{hybrik} & 41.8 & 71.6 & 82.3 & -- & 65.6 & 103.0 & 122.2 & -- & 56.4 & 96.8 & 110.4 & -- \\
            & HMR2.0~\cite{hmr2} & 44.4 & 69.8 & 82.2 & \textbf{18.1} & 60.6 & 98.0 & 120.3 & \textbf{19.8} & 48.1 & 96.0 & 110.9 & \underline{18.8} \\
            & ReFit$^*$~\cite{refit} & 40.5 & 65.3 & 75.1 & \underline{18.5} & 58.6 & 88.0 & 104.5 & \underline{20.7} & \underline{47.9} & \underline{80.7} & \underline{92.9} & \textbf{17.1} \\
            & TokenHMR~\cite{dwivedi2024tokenhmradvancinghumanmesh} & 44.3 & 71.0 & 84.6 & -- & 55.6 & 91.7 & 109.4 & -- & -- & -- & -- & --\\
            & CameraHMR~\cite{camerahmr}& \underline{38.5} & \underline{62.1} & \underline{72.9} & {--} & \underline{43.7} & \underline{73.0} & \underline{85.4} & {--}& {--} & {--} & {--} & {--}    \\
            & PromptHMR$^\dagger$~\cite{wang2025prompthmrpromptablehumanmesh}& \textbf{36.6} & \textbf{58.7} & \textbf{69.4} & {--} & \textbf{41.0} & \textbf{71.7} & \textbf{84.5} & {--}& \textbf{37.3} & \textbf{56.6} & \textbf{65.5} & {--} \\

            \cmidrule{1-14}

            \multirow{12}{1em}{\rotatebox[origin=c]{90}{\textbf{temporal}}}

            & TCMR$^*$~\cite{tcmr} & 52.7 & 86.5 & 101.4 & 6.0 & 79.6 & 127.6 & 147.9 & 5.3 & 65.6 & 119.1 & 137.7 & 5.0 \\
            & VIBE$^*$~\cite{vibe} & 51.9 & 82.9 & 98.4 & 18.5 & 81.4 & 125.9 & 146.8 & 26.6 & 68.4 & 120.5 & 140.2 & 21.8 \\
            & MPS-Net$^*$~\cite{mpsnet} & 52.1 & 84.3 & 99.0 & 6.5 & 81.3 & 123.1 & 138.4 & 6.2 & 67.1 & 118.2 & 136.7 & 5.8 \\
            & GLoT$^*$~\cite{glot} & 50.6 & 80.7 & 96.4 & 6.0 & 78.8 & 119.7 & 138.4 & 5.4 & 65.6 & 114.3 & 132.7 & 5.2 \\
            & GLAMR~\cite{glamr} & 51.1 & -- & -- & 8.0 & 73.5 & 113.6 & 133.4 & 32.9 & 79.9 & -- & -- & 107.7 \\
            & TRACE$^*$~\cite{trace} & 50.9 & 79.1 & 95.4 & 28.6 & 70.9 & 109.9 & 127.4 & 25.5 & -- & -- & -- & -- \\
            & SLAHMR~\cite{slahmr} & 55.9 & -- & -- & -- & 69.5 & 93.5 & 110.7 & 7.1 & 52.5 & -- & -- & 9.4 \\
            & PACE~\cite{pace} & -- & -- & -- & -- & -- & -- & -- & -- & 49.3 & -- & -- & 8.8 \\
            & WHAM$^*$~\cite{wham} & 35.9 & 57.8 & 68.7 & 6.6 & 50.4 & 79.7 & 94.4 & 5.3 & 44.3 & 80.0 & 91.2 & 5.3 \\
            & TRAM$^*$~\cite{tram} & \underline{35.6} & 59.3 & 69.6 & \underline{4.9} & 45.7 & 74.4 & \underline{86.6} & 4.9 & -- & -- & -- & -- \\
            & GVHMR$^*$~\cite{gvhmr} & 36.2 & \underline{55.6} & \underline{67.2} & 5.0 & \underline{42.7} & \underline{72.6} & \textbf{84.2} & \underline{3.6} & \underline{39.8} & \underline{66.1} & \underline{74.5} & \underline{4.1} \\
            & Multi-HMR~\cite{multihmr} & 45.9 & 73.1 & 87.1 & -- & 50.1 & 81.6 & 95.7 & -- & -- & -- & -- & --\\
            & MaQ$^*$~\cite{liu2025MoaQ} & 44.7 & 72.6 & 84.9 & -- & -- & -- & -- & -- & -- & -- & -- & -- \\
            & Human3R~\cite{human3r} & 44.1 & 71.2 & 84.9 & -- & 48.5 & 73.9 & 86.0 & -- & -- & -- & -- & -- \\

            \cmidrule{2-14} 

            & WATCH$^*$ (Ours)& \textbf{35.4} & \textbf{54.5} & \textbf{66.3} & \textbf{4.7} & \textbf{42.4} & \textbf{72.5} & \underline{84.3} & \textbf{3.1} & \textbf{36.6} & \textbf{64.6} & \textbf{73.6} & \textbf{3.6} \\

            \cmidrule[0.75pt]{1-14}
        \end{tabular}
    }}
    \caption{
    \textbf{Camera-space metrics.} Parentheses denote joint count. Bold/underline mark numerically best/second values within each block under each method's published protocol, \rev{not a common-protocol ranking}. $^*$ denotes 3DPW training; \rev{$^\dagger$ uses GT focal length}. WATCH uses flip-test averaging.
    }
    \label{tab:quant_camera}

\end{table*}

\subsection{Implementation Details}
We trained WATCH on a composite dataset consisting of AMASS~\cite{amass}, BEDLAM~\cite{bedlam}, BEDLAM2~\cite{bedlam2}, H36M~\cite{h36m}, and 3DPW~\cite{3dpw}. The model was trained for 12 hours on two RTX 4090 GPUs with a sequence length of 120 and a batch size of 128, converging within 500 epochs. At inference, WATCH runs at approximately 30 FPS on a single RTX 4090, comparable to GVHMR, with a similar model parameter count.

Following WHAM and GVHMR, we simulate mixed static and dynamic camera trajectories on AMASS, generate bounding boxes, normalize keypoints to [-1, 1], and set image features to zero. For other video datasets, image features are extracted using a frozen HMR2.0 encoder~\cite{hmr2}. At inference, we use DPVO~\cite{dpvo} for in-the-wild videos or GT camera extrinsics where available (e.g., EMDB).

\noindent\textbf{Velocity Processing.} We normalize camera angular velocity with training-set statistics. Local camera translation is passed directly through its modality projector; it is not divided by bounding-box height. Because monocular VO translation remains scale-ambiguous, WATCH uses it only as a soft learned condition and never hard-decodes it as metric human displacement.
\subsection{Datasets and Evaluation Metrics}
\noindent\textbf{Datasets. }
Following~\cite{wham,gvhmr}, we evaluate on 3DPW~\cite{3dpw}, RICH~\cite{rich}, and EMDB~\cite{emdb}. For world-space evaluation we use RICH (191 sequences, static camera) and EMDB-2 (25 sequences, dynamic hand-held camera with GT IMU). For camera-space evaluation we additionally use EMDB-1 (17 sequences) and 3DPW (37 sequences, in-the-wild moving camera). The two world-space benchmarks represent opposite ends of the camera-motion spectrum: RICH tests robustness when no camera motion signal is available; EMDB-2 tests whether camera translation from SLAM is effectively leveraged.

\vspace{1mm}
\noindent\textbf{Metrics. }
Following WHAM~\cite{wham}, we split sequences into non-overlapping 100-frame windows. WA-MPJPE$_{100}$ fits a similarity transform over each complete window; W-MPJPE$_{100}$ fits it on the first two frames and evaluates the rest, exposing within-window drift~\cite{slahmr, pace}. Tables aggregate all valid frames by micro averaging. RTE measures normalized root-trajectory error, Jitter measures acceleration noise, and Foot-Sliding measures contact artifacts. Camera-space metrics are PA-MPJPE, MPJPE, PVE, and Acceleration Error.
\subsection{World-Space Motion Evaluation}

Table~\ref{tab:quant_global} compares WATCH against state-of-the-art methods on RICH and EMDB-2. Global HMR methods fall into two regimes with fundamentally different operating points: \textbf{camera-trajectory-centric} methods directly decode SLAM trajectories, achieving low error on dynamic-camera benchmarks but assuming reliable camera translation; \textbf{human-motion-centric} methods estimate motion from visual cues, trading peak dynamic-camera accuracy for robustness across camera conditions. WATCH belongs to the second regime and is designed to remain competitive across both static and dynamic cameras without assuming SLAM reliability.

On RICH, WATCH attains WA-MPJPE$_{100}$ of 66.8mm, compared with 78.8mm for GVHMR; Jitter drops from 12.8 to 5.3 and Foot-Sliding from 3.0 to 2.2. On EMDB-2 with GT camera, WATCH obtains 103.3/251.6mm WA/W-MPJPE$_{100}$, compared with 109.1/274.9mm for GVHMR under the same camera source. With estimated DPVO input, WATCH obtains 105.2/256.9mm. These camera-source-matched comparisons show that the model remains effective with estimated input; we do not attribute this behavior to metric scale recovery.

PromptHMR-vid is stronger on EMDB-2 trajectory accuracy \rev{(71.0/216.5mm WA/W-MPJPE$_{100}$ and 1.3 RTE)}, while WATCH is smoother \rev{(10.4 Jitter and 3.1 Foot-Sliding versus 16.3 and 3.5)}, is additionally evaluated on static-camera RICH, and remains competitive with estimated-camera input. Camera-trajectory-centric methods are preferable when dynamic-camera SLAM is reliable and metric trajectory is the sole priority; WATCH targets static or uncertain camera conditions where smoothness and physical plausibility also matter. \begin{figure}[htbp]
    \centering
    \includegraphics[width=\linewidth,height=0.60\linewidth,keepaspectratio]{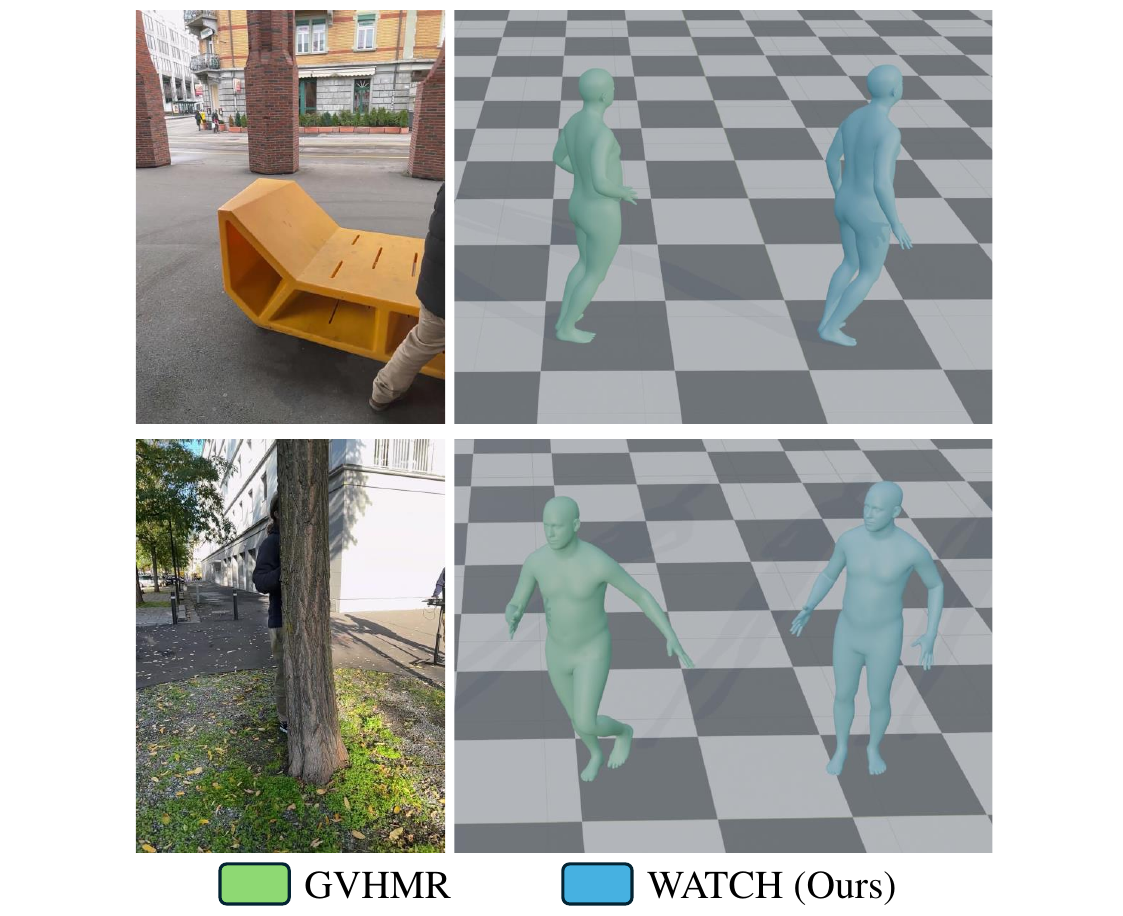}
    \vspace{-2mm}
    \caption{Robustness under partial occlusion. GVHMR produces physically implausible poses when the subject is occluded: excessive forward lean under lower-body occlusion~(top) and an unnatural deep-squat posture when the body is partially hidden by a tree~(bottom). WATCH maintains plausible upright configurations by relying on learned motion priors rather than accumulating orientation errors from ambiguous visual evidence.}
    \label{fig:occlusion}
\end{figure}

\subsection{Camera-Space Motion Evaluation}

Table~\ref{tab:quant_camera} presents our camera-space performance across the 3DPW, EMDB, and RICH datasets. Methods differ in image features and published evaluation protocols: WHAM, GVHMR, and WATCH use ViT features from HMR2.0, whereas PromptHMR uses DINOv2~\cite{oquab2023dinov2} with multimodal supervision and GT focal length, as marked in the table.

On 3DPW, WATCH attains PA-MPJPE of 35.4mm and MPJPE of 54.5mm, compared with 36.2/55.6mm for GVHMR, with lower acceleration error (4.7 vs.\ 5.0). On EMDB-1, WATCH obtains 42.4mm PA-MPJPE and 3.1 acceleration error (GVHMR: 42.7 and 3.6). On RICH, WATCH improves GVHMR by 3.2mm PA-MPJPE and 1.5mm MPJPE. Under PromptHMR's published protocol, PromptHMR is stronger than WATCH on EMDB-1 PA-MPJPE/MPJPE \rev{(41.0/71.7mm vs.\ 42.4/72.5mm)} and RICH MPJPE/PVE \rev{(56.6/65.5mm vs.\ 64.6/73.6mm)}.

\begin{figure}[htbp]
    \centering
    \includegraphics[width=\linewidth]{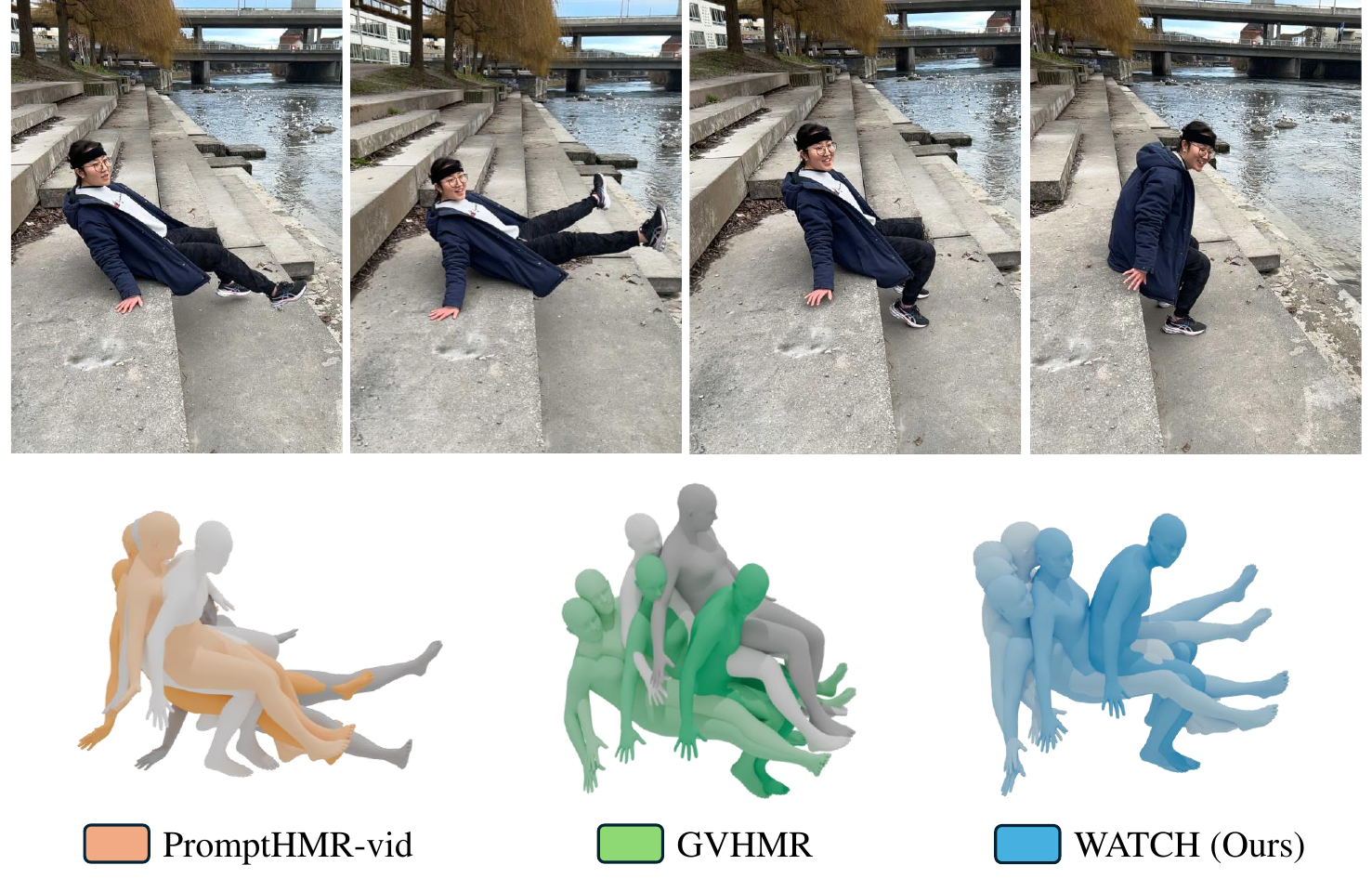}
    \caption{Qualitative comparison with PromptHMR-vid~\cite{wang2025prompthmrpromptablehumanmesh} and GVHMR~\cite{gvhmr} on global human motion estimation. Frames highlighted in gray indicate cases where baseline methods produce physically implausible poses or severe artifacts, while WATCH maintains consistent and plausible reconstruction.}
    \label{fig:result}
\end{figure}

\begin{figure}[htbp]
    \centering
    \includegraphics[width=\linewidth]{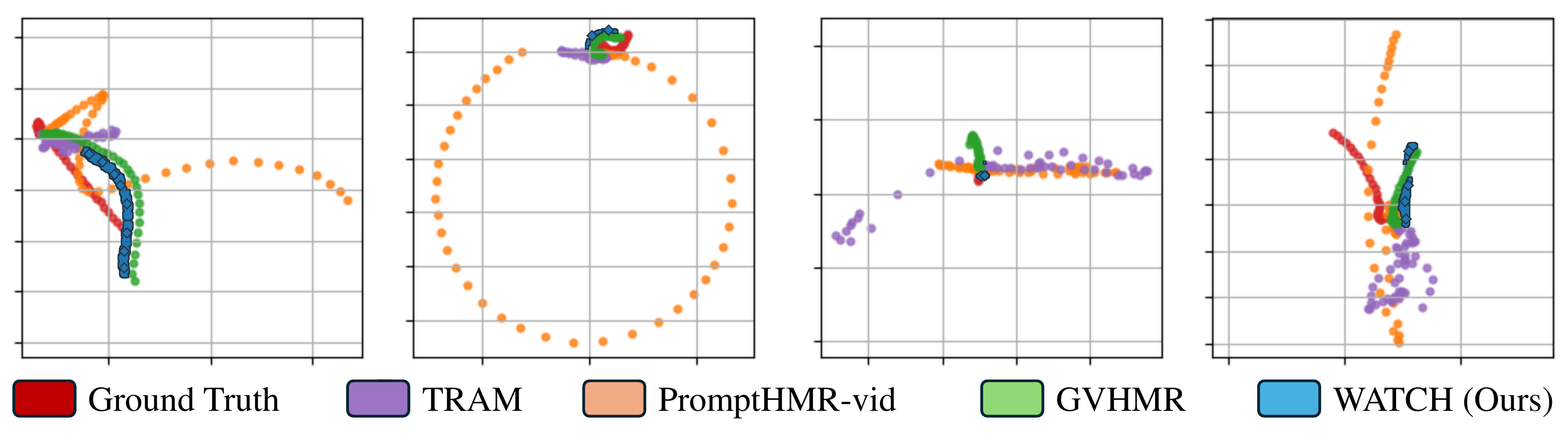}
    \caption{Bird's-eye trajectories on EMDB. WATCH uses blue diamonds with a dark outline; other methods use circular markers and distinct colors, making overlapping traces visible. Red denotes GT.}
    \label{fig:traj}
\end{figure}

\begin{table}[tbh]
    \centering
    \setlength{\tabcolsep}{4pt}
    \renewcommand{\arraystretch}{1.2}
    \resizebox{\columnwidth}{!}{
        \begin{tabular}{l|cccc|ccc}
            \toprule[0.75pt]
            \multicolumn{1}{c}{} & \multicolumn{4}{c|}{\textbf{EMDB-2 (World-space)}} & \multicolumn{3}{c}{\textbf{EMDB-1 (Camera-space)}} \\
            \cmidrule(lr){2-5} \cmidrule(lr){6-8}
            \textbf{Method Variant} & \scriptsize{WA-MPJPE$_{100}$} & \scriptsize{W-MPJPE$_{100}$} & \scriptsize{RTE} & \scriptsize{FS} & \scriptsize{PA-MPJPE} & \scriptsize{MPJPE} & \scriptsize{PVE} \\
            \midrule

            \textbf{w/ DPVO} \\
            \hspace{1em}WHAM & 135.6 & 354.8 & 6.0 & 4.4 & 50.4 & 79.7 & 94.4 \\
            \hspace{1em}GVHMR & \underline{111.0} & \underline{276.5} & \underline{2.0} & 3.5 & \underline{42.7} & 72.6 & \underline{84.2} \\
            \hspace{1em}WATCH (w/o cam traj.) & 112.0 & 281.0 & 2.1 & \underline{3.3} & \underline{42.7} & \textbf{72.1} & \textbf{83.3} \\
            \hspace{1em}WATCH (w/ cam traj.) & \textbf{105.2} & \textbf{256.9} & \textbf{1.9} & \textbf{3.1} & \textbf{42.4} & \underline{72.5} & 84.3 \\
            \addlinespace
            \midrule

            \textbf{w/ GT cam} \\
            \hspace{1em}WHAM & 131.1 & 335.3 & 4.1 & 4.4 & 50.4 & 79.7 & 94.4 \\
            \hspace{1em}GVHMR & \underline{109.1} & \underline{274.9} & \underline{1.9} & 3.5 & \underline{42.7} & 72.6 & \underline{84.2} \\
            \hspace{1em}WATCH (w/o cam traj.) & 111.0 & 278.0 & \underline{1.9} & \underline{3.2} & \underline{42.7} & \textbf{72.1} & \textbf{83.3} \\
            \hspace{1em}WATCH (w/ cam traj.) & \textbf{103.3} & \textbf{251.6} & \textbf{1.7} & \textbf{3.1} & \textbf{42.4} & \underline{72.5} & 84.3 \\
            \bottomrule[0.75pt]
        \end{tabular}
    }
    \caption{\textbf{Ablation study: effect of camera trajectory conditioning.} EMDB-2 metrics are in world space; EMDB-1 in camera space. \textbf{WATCH (w/o cam traj.)} is trained with the camera velocity input removed, isolating the benefit of trajectory conditioning. \textbf{WATCH (w/ cam traj.)} is the full method. Adding camera trajectory reduces WA-MPJPE$_{100}$ by 6.8mm~(DPVO) / 7.7mm~(GT cam) and Foot-Sliding from 3.2--3.3 to 3.1 across both camera sources, while camera-space metrics remain on par.}
    \label{tab:quant_emdb_world_camera_combined}
\end{table}

\subsection{Ablation Study}
\label{sec:ablation}
Table~\ref{tab:exact_yaw} applies exact yaw recovery to GVHMR at inference time. In this paired matched reproduction, replacing only world-heading recovery reduces WA-MPJPE$_{100}$ by 6.8mm on RICH and 2.9mm on EMDB-2 with GT camera, and Jitter by 2.7 and 2.3, respectively, without retraining.

\begin{table}[t]
    \centering
    \setlength{\tabcolsep}{3pt}
    \renewcommand{\arraystretch}{1.1}
    \resizebox{\columnwidth}{!}{
        \begin{tabular}{l?cccc?cccc}
            \cmidrule[0.75pt]{1-9}
            & \multicolumn{4}{c?}{\textbf{RICH}} & \multicolumn{4}{c}{\textbf{EMDB-2}} \\
            \cmidrule(lr){2-5} \cmidrule(lr){6-9}
            \textbf{Method} & \scriptsize{WA-MPJPE$_{100}$}$\downarrow$ & \scriptsize{W-MPJPE$_{100}$}$\downarrow$ & \scriptsize{RTE}$\downarrow$ & \scriptsize{Jitter}$\downarrow$ & \scriptsize{WA-MPJPE$_{100}$}$\downarrow$ & \scriptsize{W-MPJPE$_{100}$}$\downarrow$ & \scriptsize{RTE}$\downarrow$ & \scriptsize{Jitter}$\downarrow$ \\
            \cmidrule{1-9}
            \textbf{EMDB-2: DPVO; RICH: provided static cam} \\
            \hspace{1em}GVHMR re-eval.~\cite{gvhmr} & 78.8 & 126.3 & 2.4 & 12.8 & 109.5 & 274.3 & \underline{2.0} & 16.3 \\
            \hspace{1em}~~~+~Exact Yaw~(Ours)       & \underline{72.0} & \underline{114.9} & \underline{2.3} & \underline{10.1} & \underline{107.5} & \underline{268.7} & \textbf{1.9} & \underline{14.2} \\
            \cmidrule(lr){1-9}
            \hspace{1em}WATCH~(Ours)                & \textbf{66.8} & \textbf{107.9} & \textbf{2.1} & \textbf{5.3} & \textbf{105.2} & \textbf{256.9} & \textbf{1.9} & \textbf{10.4} \\
            \cmidrule{1-9}
            \textbf{EMDB-2: GT cam; RICH: provided static cam} \\
            \hspace{1em}GVHMR re-eval.~\cite{gvhmr} & 78.8 & 126.3 & 2.4 & 12.8 & 109.1 & 272.8 & \underline{1.9} & 16.1 \\
            \hspace{1em}~~~+~Exact Yaw~(Ours)       & \underline{72.0} & \underline{114.9} & \underline{2.3} & \underline{10.1} & \underline{106.2} & \underline{265.3} & \textbf{1.7} & \underline{13.8} \\
            \cmidrule(lr){1-9}
            \hspace{1em}WATCH~(Ours)                & \textbf{66.8} & \textbf{107.9} & \textbf{2.1} & \textbf{5.3} & \textbf{103.3} & \textbf{251.6} & \textbf{1.7} & \textbf{9.8} \\
            \cmidrule[0.75pt]{1-9}
        \end{tabular}
    }
    \caption{\textbf{Effect of analytical heading recovery.} In each block, the first two rows rerun the same public GVHMR checkpoint with one evaluator; only world-heading recovery is replaced, without retraining. These paired re-evaluation values are reported separately from the GVHMR reference entries in Table~\ref{tab:quant_emdb_world_camera_combined}; only within-block differences are interpreted. WATCH uses the submitted e459 checkpoint. RICH has one provided-static-camera evaluation, repeated across blocks only for side-by-side EMDB-2 comparison.}
    \label{tab:exact_yaw}
\end{table}

\begin{table}[t]
    \centering
    \setlength{\tabcolsep}{3.2pt}
    \renewcommand{\arraystretch}{1.0}
    \resizebox{\columnwidth}{!}{
        \begin{tabular}{lccccc}
            \toprule
            \textbf{Controlled variant} & \textbf{WA} & \textbf{W} & \textbf{RTE} & \textbf{Jitter} & \textbf{FS} \\
            \midrule
            Control reference ($\lambda_h=1$) & 104.8 & 256.3 & 1.6 & 9.8 & 3.1 \\
            w/o explicit camera-output losses & 106.0 & 264.4 & 1.7 & 9.8 & 3.0 \\
            Direct full-camera rotation & 132.4 & 386.2 & 10.7 & 18.6 & 3.7 \\
            $\lambda_h=0$ & 105.3 & 260.5 & 1.7 & 10.0 & 3.3 \\
            $\lambda_h=0.5$ & 103.6 & 253.9 & 1.7 & 9.9 & 3.1 \\
            \midrule
            Full WATCH (Ours) & \textbf{103.3} & \textbf{251.6} & 1.7 & 9.8 & 3.1 \\
            \bottomrule
        \end{tabular}
    }
    \caption{Controlled ablations on EMDB-2 with GT camera. Absolute metrics are reported (WA/W in mm). The first five rows use one separately trained control recipe. The separated Full WATCH (Ours) row reports the submitted checkpoint from Table~\ref{tab:quant_emdb_world_camera_combined} for context only and is not used for cross-recipe ablation comparisons; protocol details are in the supplement.}
    \label{tab:controlled_ablation}
    \vspace{-1.5mm}
\end{table}

\rev{Table~\ref{tab:controlled_ablation} reports absolute metrics for separately trained controls under one common recipe; its $\lambda_h=1$ reference obtains 104.8/256.3mm WA/W-MPJPE$_{100}$. The submitted Full checkpoint is shown separately for context at 103.3/251.6mm and is not used for cross-recipe ablation comparisons. Without explicit camera-output losses, WA/W is 106.0/264.4mm; direct full-camera-rotation prediction obtains 132.4/386.2mm and 18.6 Jitter. For $\lambda_h=0$ and $0.5$, WA/W is 105.3/260.5mm and 103.6/253.9mm, respectively. These controls isolate auxiliary camera supervision, orientation formulation, and human-trajectory consistency; complete protocol details are provided in the supplement.}

Table~\ref{tab:quant_emdb_world_camera_combined} isolates the camera-translation input: removing it changes WA/W-MPJPE$_{100}$ from 105.2/256.9 to 112.0/281.0mm with DPVO and from 103.3/251.6 to 111.0/278.0mm with GT camera, gains of 6.8/24.1mm and 7.7/26.4mm.

On all 25 EMDB-2 sequences with GT camera and first-heading alignment, WATCH obtains $0.662^\circ$ yaw MAE and $0.303^\circ$ sliding 100-frame drift, versus $2.428^\circ$ and $0.766^\circ$ when only the recovery rule is changed to GVHMR-style rollout. This reduces MAE by 72.7\% and drift by approximately 60\%.

The inference-time replacement requires relative camera rotation, a compatible roll-pitch/gravity state, and explicit world-heading reconstruction. GVHMR exposes these interfaces; WHAM does not expose all three and requires method-specific modification and retraining.

On all 25 EMDB-2 sequences, oracle-box-prompted SAM~\cite{sam} with the same DROID-SLAM~\cite{droid}, ZoeDepth~\cite{zoedepth}, checkpoint, and evaluator improves WA/W-MPJPE$_{100}$ from 113.1/277.6 to 106.2/260.5mm. Masking improves the camera input and is compatible with WATCH, but cannot determine human root depth or metric scale, nor recover human world motion under a static camera; it does not replace world-motion reconstruction.

\noindent\textbf{Effect on static-camera scenes.} On RICH, camera translation is near zero; Table~\ref{tab:exact_yaw} therefore isolates the benefit of analytical heading recovery from translation conditioning.

\noindent\textbf{Running time.} WATCH performs reconstruction in a single feed-forward pass without iterative bundle adjustment or dense scene reconstruction.

\subsection{Limitations}
WATCH relies on camera extrinsics and may degrade when VO fails in textureless or heavily occluded scenes. Long-sequence integration can accumulate error~\cite{wham,gvhmr}. Joint camera-scene-human methods offer richer scene constraints through offline optimization or dense reconstruction, whereas WATCH targets lightweight feed-forward inference. The settings are complementary; explicit scene constraints and multi-person reconstruction remain future work.


\section{Conclusion}

We present WATCH, which improves global human motion reconstruction by addressing two signals that prior human-motion-centric methods leave unused: camera orientation is processed through a fixed transformation with no independent supervision, and camera translation is discarded entirely. We propose an analytical heading decomposition that supervises roll-pitch and yaw separately, and a soft camera trajectory conditioning mechanism that incorporates camera velocity as a learned prior without hard-decoding. These designs achieve consistent gains on both static-camera (RICH) and dynamic-camera (EMDB) benchmarks, demonstrating that human-motion-centric methods can effectively leverage camera translation while preserving plausible motion.

\clearpage
\bibliographystyle{eg-alpha-doi}
\bibliography{main}


\end{document}


\maketitle
\vspace{-2em}

\section{Heading Recovery Details}
\subsection{Coordinate convention}
We write $\mathbf R_t^{cam}$ for camera-to-world rotation and use the
body-frame increment
\begin{equation}
 \Delta\mathbf R_t^{cam}=(\mathbf R_t^{cam})^T\mathbf R_{t+1}^{cam}.
\end{equation}
The preprocessing code stores the equivalent world-to-camera increment, which
is transposed before heading recovery. This convention avoids an inverse yaw
increment when reproducing the implementation.

We factor
$\mathbf R_t^{cam}=\mathbf R_{yaw,t}^{cam}\mathbf R_{rp,t}^{cam}$, where
$\mathbf R_{yaw,t}^{cam}$ rotates only about the gravity axis. The relative
heading is
\begin{equation}
 \Delta\mathbf R_{yaw,t}^{cam}=
 \mathbf R_{rp,t}^{cam}\Delta\mathbf R_t^{cam}
 (\mathbf R_{rp,t+1}^{cam})^T,
\end{equation}
and is integrated as
\begin{equation}
 \mathbf R_{yaw,t+1}^{cam}=\mathbf R_{yaw,t}^{cam}
 \Delta\mathbf R_{yaw,t}^{cam}.
\end{equation}
Only relative heading affects the reconstructed motion, so the first heading
can be set to identity at inference.

\subsection{Numerically stable factorization}
In the Y-down convention, the camera forward vector $\mathbf f$ is projected
to the horizontal plane,
\begin{equation}
 \mathbf f_{xz}=\mathbf f-(\mathbf f^T\mathbf e_y)\mathbf e_y.
\end{equation}
When $\|\mathbf f_{xz}\|>10^{-6}$ we normalize it; otherwise we use a fixed
horizontal basis vector. Cross products reconstruct an orthonormal heading
basis. This handles cameras looking nearly parallel to gravity.

\section{Architecture and Inputs}
Each frame token starts from 17 confidence-weighted 2D joints. Independent
projectors map HMR2.0 image features, CLIFF bounding-box/intrinsics features,
normalized camera angular velocity, and local camera-translation velocity to
512 dimensions. The latter is passed directly to its projector and is
\emph{not} divided by bounding-box height. The projected conditions are added,
preserving one token per frame.

The backbone is a 12-layer, 512-dimensional, 8-head RoPE Transformer with MLP
ratio 4 and training sequence length 120. Output heads predict 142 HMR values, 3
weak-perspective camera values, 15-dimensional human motion,
15-dimensional camera motion, and six contact logits. An endpoint head with
512 input and 256 hidden dimensions predicts a sequence-level residual that is
used to fuse forward and backward human-trajectory rollouts.

Human and camera trajectories use distinct states:
\begin{equation}
 \hat{\mathbf t}_t^q=\mathbf t_0^q+
 \sum_{i=1}^{t}\mathbf R_i^q\hat{\mathbf v}_i^q,
 \qquad q\in\{h,cam\}.
\end{equation}

\section{Implementation-Authoritative Objective}
The submitted Full WATCH checkpoint and the corresponding Full WATCH results use
the following implementation-authoritative objective:
\begin{equation}
\begin{split}
 \mathcal L={}&\mathcal L_{simple}+\mathcal L_{hmr}
 +\mathcal L_{traj}^{h}+\mathcal L_{traj}^{cam}\\
 &+\mathcal L_{static}+\lambda_{accel}\mathcal L_{accel}.
\end{split}
\end{equation}
Here $\lambda_{accel}=30$.
The L2 trajectory notation and $\lambda_{cam}=0.25$ in the submitted main paper,
and the omission of $\mathcal L_{static}$ and $\mathcal L_{accel}$ from the
submitted supplement, were documentation errors rather than different
Full-model settings.
$\mathcal L_{simple}$ is MSE on encoded HMR, human-motion, and camera-motion
outputs. $\mathcal L_{hmr}$ weights root-relative 3D joints and vertices by
500 each, 2D joints and vertices by 1000 each, and weak-perspective camera
parameters by 1.

For the human branch, the two teacher-forced rollouts are
\begin{align}
 \hat{\mathbf t}^{(1)}_t &= \mathbf t_0+
 \sum_{i=1}^{t}\mathbf R_{gt,i}^{h}\hat{\mathbf v}_i^h,\\
 \hat{\mathbf t}^{(2)}_t &= \mathbf t_0+
 \sum_{i=1}^{t}\hat{\mathbf R}_i^{h}\mathbf v_{gt,i}^h.
\end{align}
Their L1 errors form $\mathcal L_{traj}^{h}$; the camera branch is identical.
Both trajectory terms have unit weight. $\mathcal L_{static}$ is unit-weight
BCE on six automatically labelled contact logits. The acceleration term is
componentwise smooth-L1 with $\beta=0.02$:
\begin{align}
 \mathbf a_t &={}
 \hat{\mathbf J}^{c,rel}_{t-1}-2\hat{\mathbf J}^{c,rel}_{t}
 +\hat{\mathbf J}^{c,rel}_{t+1},\nonumber\\
 \mathcal L_{accel} &={}
 \operatorname*{mean}_{t,j,k\,\mathrm{valid}}\rho_{\beta}(a_{t,j,k}).
\end{align}
\begin{equation}
 \rho_{\beta}(x)=
 \begin{cases}
 x^2/(2\beta), & |x|<\beta,\\
 |x|-\beta/2, & \text{otherwise},
 \end{cases}
\end{equation}
The mean is over valid frame triplets, joints, and Cartesian components.

\section{Training and Evaluation Protocol}
The submitted model was trained for 500 epochs on two RTX 4090 GPUs with
sequence length 120 and batch size 128. Training took approximately 12 hours. Submitted numbers use
checkpoint e459-s266800, horizontal-flip test-time averaging, and postprocessing
version 1.

World-space evaluation splits sequences into non-overlapping 100-frame
windows. WA-MPJPE$_{100}$ fits a similarity transform over a complete window;
W-MPJPE$_{100}$ fits it using the first two frames and evaluates the remaining
frames. Reported values micro-average all valid frames.

\section{Additional Controlled Evaluations}
\begin{table}[t]
\centering
\small
\setlength{\tabcolsep}{2.0pt}
\begin{tabular}{lccccc}
\toprule
Variant & WA & W & RTE & Jit. & FS\\
\midrule
Control reference & 104.8 & 256.3 & 1.6 & 9.8 & 3.1\\
No explicit cam-output losses & 106.0 & 264.4 & 1.7 & 9.8 & 3.0\\
Direct full-camera rotation & 132.4 & 386.2 & 10.7 & 18.6 & 3.7\\
$\lambda_h=0$ & 105.3 & 260.5 & 1.7 & 10.0 & 3.3\\
$\lambda_h=0.5$ & 103.6 & 253.9 & 1.7 & 9.9 & 3.1\\
\midrule
Full WATCH (Ours) & \textbf{103.3} & \textbf{251.6} & 1.7 & 9.8 & 3.1\\
\bottomrule
\end{tabular}
\caption{Matched EMDB-2 controls with GT camera (WA/W in mm). The first five rows
use the same 500-epoch, four-GPU, 192-sequences/GPU retraining recipe, the final
e499-s096500 checkpoint of each run, and flip averaging. The separated Full
WATCH (Ours) row reports submitted e459 only for context and is not part of matched
control comparisons. The 104.8/256.3 reference is used only among the separately
trained controls; submitted Full remains 103.3/251.6mm.}
\label{tab:controls-supp}
\end{table}

The no-explicit-camera-output-losses variant retains camera-translation input
but sets both direct and trajectory camera-output loss weights to zero; other
world-space losses are unchanged. Direct full-camera rotation
replaces roll-pitch prediction plus analytical recovery with continuous 6D
full-rotation prediction while retaining the same inputs, backbone, data,
schedule, and other auxiliary outputs and losses. The GT target is
left-multiplied by the inverse first-frame yaw, and the 6D head predicts full
rotation directly in the same canonical gauge. During training, this gauge is
restored before downstream world-space losses; inference remains in the
canonical gauge. The six decomposition MSE channels are disabled and replaced
by a squared SO(3) geodesic-angle loss weighted by $6/172$. The results isolate
the auxiliary camera task and orientation formulation. The $\lambda_h$ rows
show a benefit from nonzero human-trajectory supervision without sensitivity to
the exact weight between 0.5 and 1.

\subsection{Direct yaw accuracy}
On all 25 EMDB-2 sequences with GT camera and first-heading alignment, the
submitted Full checkpoint obtains $0.662^\circ$ yaw MAE and $0.303^\circ$
sliding 100-frame drift. On the same checkpoint and inputs, changing only the recovery
rule to GVHMR-style rollout gives $2.428^\circ$ and $0.766^\circ$. These are
frame/window micro means; analytical recovery reduces the errors by 72.7\% and
approximately 60\%.

\subsection{Masked-SLAM diagnostic}
We use oracle-box prompts to SAM on all 25 EMDB-2 sequences. Masked and
unmasked inputs use the same DROID front end, ZoeDepth metric scaling, and
no-flip evaluator; both camera rotation and translation are passed to the
submitted Full WATCH checkpoint. Masking improves WA/W-MPJPE$_{100}$ from 113.1/277.6 to
106.2/260.5mm. This establishes that masking improves the camera front end and
is compatible with WATCH. SAM masking alone does not determine monocular human
root depth or metric scale, and cannot recover human world motion under a static camera, so
it is not a replacement for WATCH.

\section{Qualitative Results}
\begin{figure}[t]
\centering
\includegraphics[width=\linewidth]{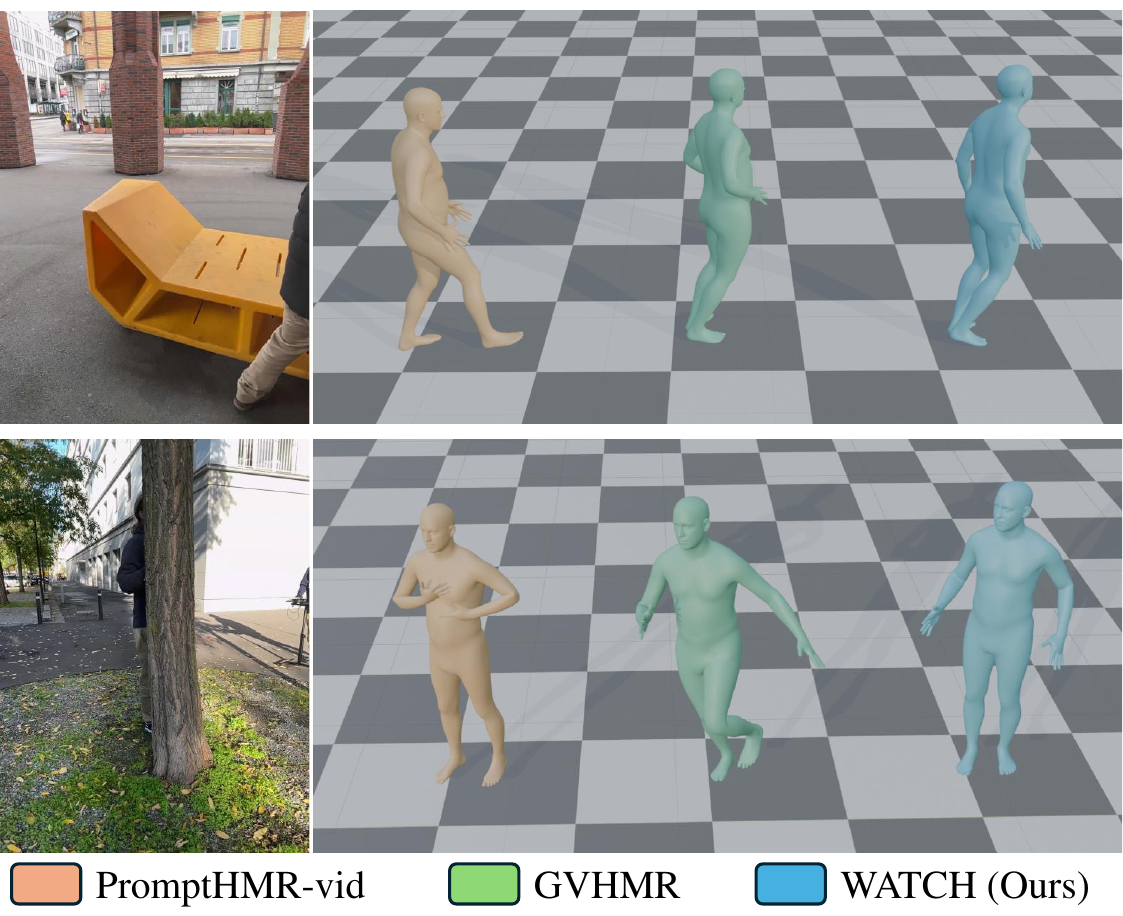}
\caption{Severe occlusions. WATCH retains plausible poses when most of the body
is hidden or outside the frame.}
\end{figure}

\begin{figure*}[t]
\centering
\includegraphics[width=0.96\textwidth]{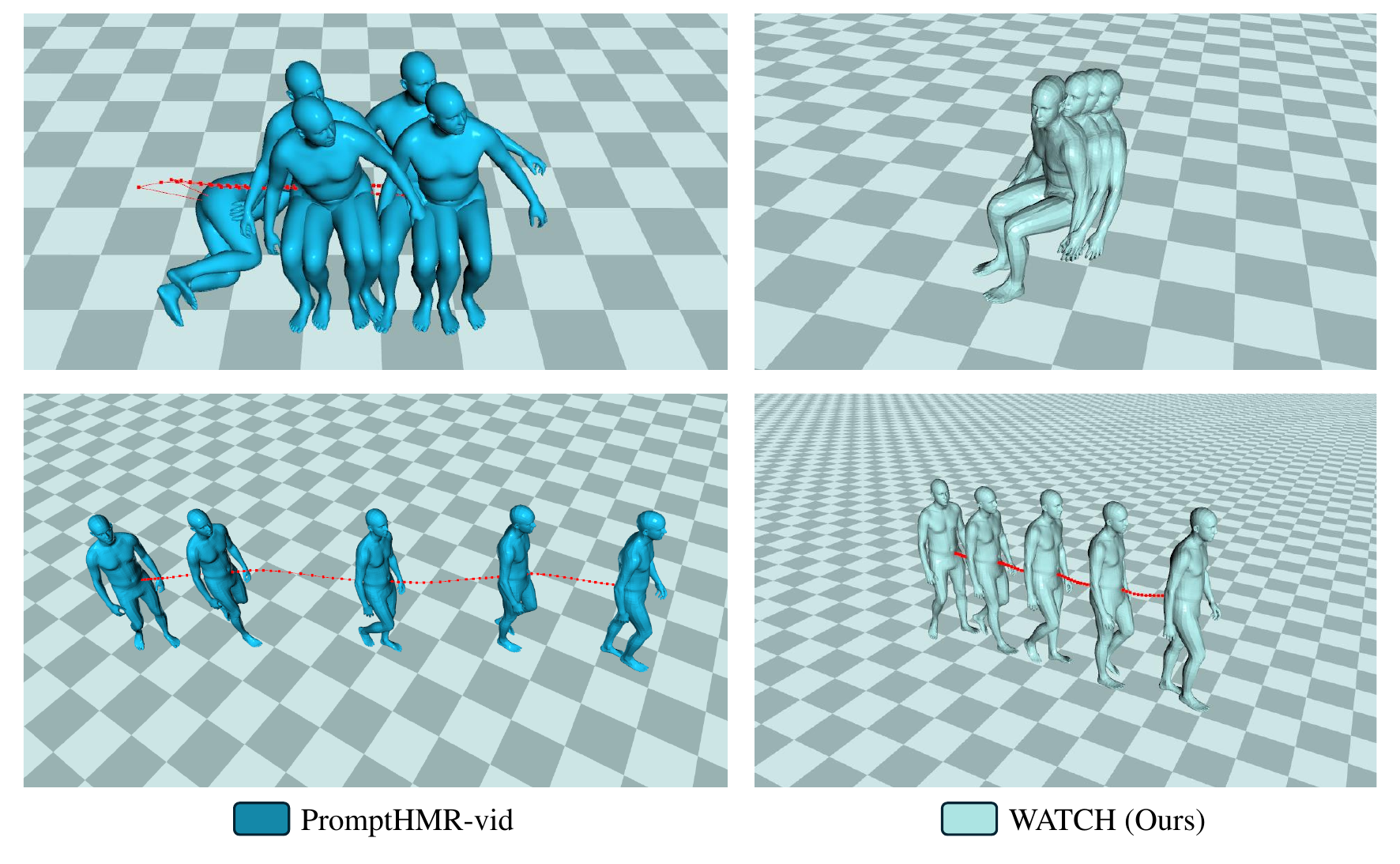}
\caption{Stationary sitting and normal walking sequences. The examples compare
long-term trajectory behavior of PromptHMR-vid and WATCH.}
\end{figure*}

\begin{figure*}[t]
\centering
\includegraphics[width=0.96\textwidth]{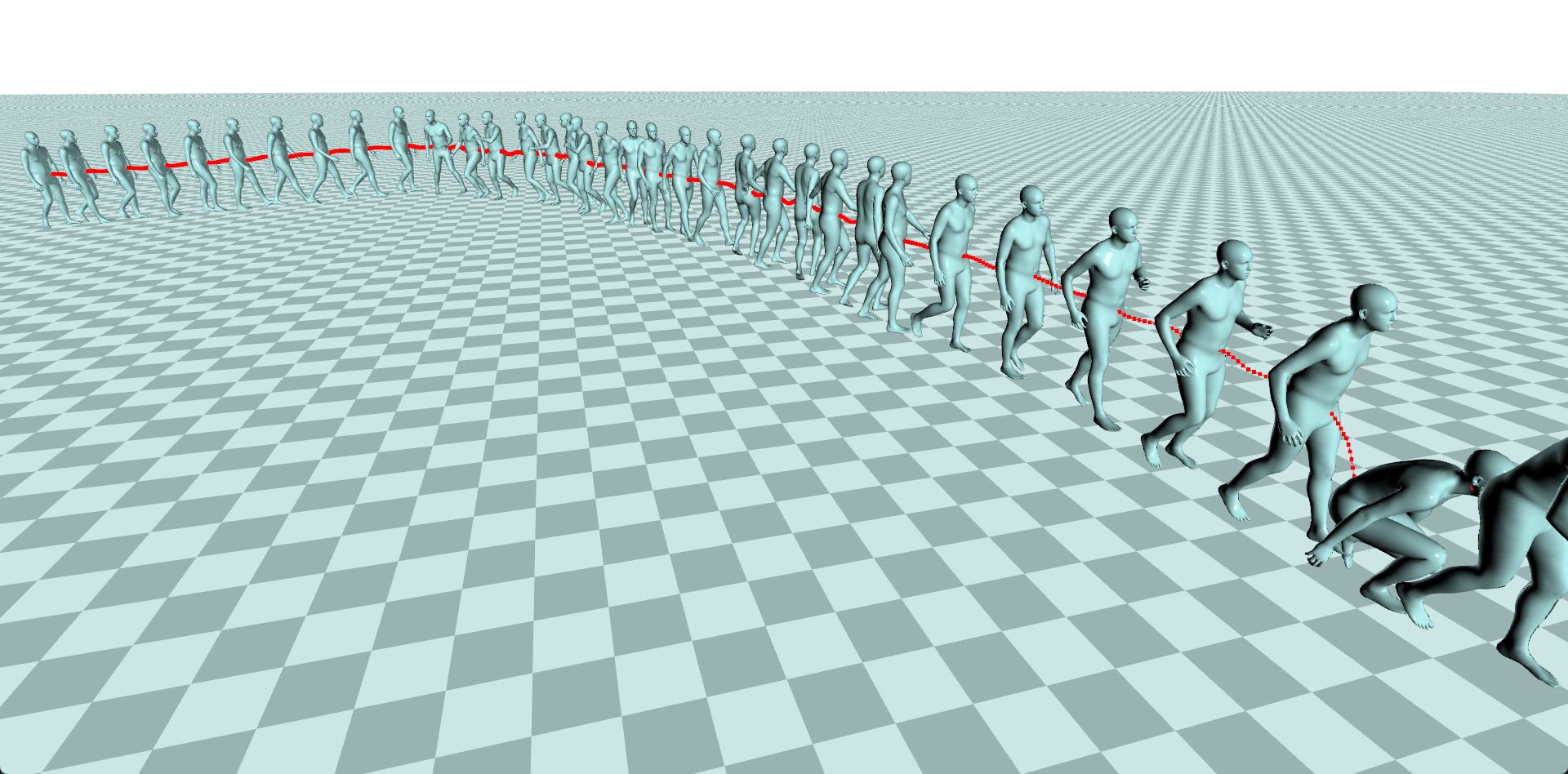}
\caption{A long sequence containing straight walking, lateral cross-stepping,
and squat-to-stand motion.}
\end{figure*}